\documentclass[graybox]{styles/svmult}

\usepackage[whole]{bxcjkjatype} 
\usepackage{hyperref}
\usepackage{amsmath}
\usepackage{tabularx}
\usepackage{array}

\usepackage{mathptmx}       
\usepackage{helvet}         
\usepackage{courier}        
\usepackage{type1cm}        
\usepackage{makeidx}         
\usepackage{graphicx}        
\usepackage{multicol}        
\usepackage{multirow}
\usepackage[bottom]{footmisc}

\makeindex             

\begin{document}

\title*{Acknowledgment~of~Emotional~States:~Generating Validating Responses for Empathetic Dialogue}
\author{Zi Haur Pang, Yahui Fu, Divesh Lala, Keiko Ochi, Koji Inoue, Tatsuya Kawahara}
\institute{Graduate School of Informatics, Kyoto University, Japan \\
\email{{pang|fu|lala|keiko|inoue|kawahara}@sap.ist.i.kyoto-u-ac.jp}}

%
%
\maketitle

\abstract*{In the realm of human-AI dialogue, the facilitation of empathetic responses is important. \textit{Validation} is one of the key communication techniques in psychology, which entails recognizing, understanding, and acknowledging others' emotional states, thoughts, and actions. This study introduces the first framework designed to engender empathetic dialogue with validating responses. Our approach incorporates a tripartite module system: 1) validation timing detection, 2) users' emotional state identification, and 3) validating response generation. Utilizing Japanese EmpatheticDialogues dataset - a textual-based dialogue dataset consisting of 8 emotional categories from Plutchik's wheel of emotions - the Task Adaptive Pre-Training (TAPT) BERT-based model outperforms both random baseline and the ChatGPT performance, in term of F1-score, in all modules. Further validation of our model's efficacy is confirmed in its application to the TUT Emotional Storytelling Corpus (TESC), a speech-based dialogue dataset, by surpassing both random baseline and the ChatGPT. This consistent performance across both textual and speech-based dialogues underscores the effectiveness of our framework in fostering empathetic human-AI communication.
}

\abstract{In the realm of human-AI dialogue, the facilitation of empathetic responses is important. \textit{Validation} is one of the key communication techniques in psychology, which entails recognizing, understanding, and acknowledging others' emotional states, thoughts, and actions. This study introduces the first framework designed to engender empathetic dialogue with validating responses. Our approach incorporates a tripartite module system: 1) validation timing detection, 2) users' emotional state identification, and 3) validating response generation. Utilizing Japanese EmpatheticDialogues dataset - a textual-based dialogue dataset consisting of 8 emotional categories from Plutchik's wheel of emotions - the Task Adaptive Pre-Training (TAPT) BERT-based model outperforms both random baseline and the ChatGPT performance, in term of F1-score, in all modules. Further validation of our model's efficacy is confirmed in its application to the TUT Emotional Storytelling Corpus (TESC), a speech-based dialogue dataset, by surpassing both random baseline and the ChatGPT. This consistent performance across both textual and speech-based dialogues underscores the effectiveness of our framework in fostering empathetic human-AI communication.
}

\section{Introduction}
\label{sec:1}

In the realm of human-robot interaction, the ability of dialogue systems to exhibit empathy is increasingly recognized as a critical component for enhancing user experience. This recognition has spurred research into developing various models that aim to infuse empathy into these systems. These models span a range of approaches, including the simulation of emotional states \cite{majumder2020mime}, the incorporation of commonsense reasoning and external knowledge sources \cite{yoo2021ep, sabour2022cem, liu2022empathetic}, and the integration of user-specific personas \cite{zhong2020towards, lin2020caire}. The effectiveness of these empathetic responses has been demonstrated in domains such as marketing \cite{liu2022artificial} and healthcare \cite{liu2018should}, where they contribute significantly to understanding human relationships and strengthening emotional connections between users and artificial agents. As such, empathetic response generation in dialogue systems not only improves the quality of interactions but also holds promise for broad application in diverse areas.

\begin{figure}[t!]
  \centering
  \includegraphics[width=6.0cm]{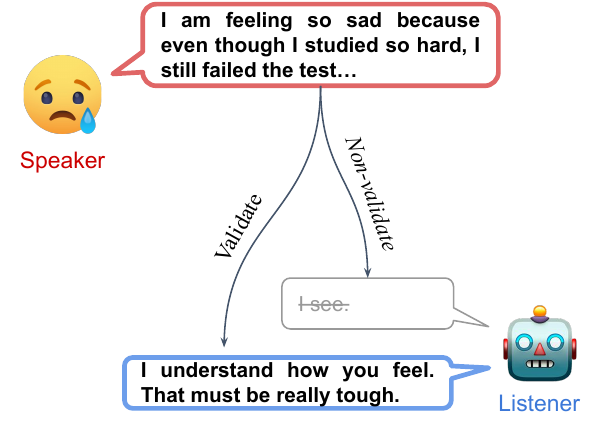} 
  \caption{Examples of dialogues with validating response and non-validating response.}
  \label{fig:example}
\end{figure}

To express empathy in dialogue system, \textit{Validation} is another communication technique used in counseling and therapy, where we recognize, understand, and acknowledge others' emotional states, thoughts, and actions. In communication, a validating statement is used to acknowledge others' feelings, showing that their emotion is being recognized and accepted. Such statements in English include ``I understand,'' ``I know exactly how you feel,'' and ``It makes sense that you feel…,'' while in Japanese including 「分かる (I understand)」,「確かにね (That's understandable)」, and「それは怖いですね (That sounds scary)」. Fig. \ref{fig:example} shows the example dialogues with validating responses and non-validating responses. 

In the domain of spoken dialogue systems, current methodologies like the Empathetic Response Generation System \cite{fuimproving} and Attentive Listening System \cite{lala2017attentive} have shown notable advancements. Nevertheless, these approaches exhibit limitations in fully addressing the emotional requirements of users, particularly in scenarios where conventional empathetic responses such as ``I am so sorry to hear that'' may not suffice. This inadequacy is especially pronounced in individuals who suppress their emotions due to stress or adverse life experiences. For such individuals, the need for acceptance and acknowledgment of their feelings - a concept known as \textit{validation} - becomes paramount. This technique has proven effective in various contexts, including chronic pain therapy, dialectical behavior therapy, and counseling \cite{edlund2015see, lambie2016role, carson2018effect}. Consequently, incorporating validation into spoken dialogue systems presents an innovative avenue for enhancing empathetic communication, catering to the specific emotional needs of this user group. This approach aligns with the findings of prior research, underscoring the significance of validation in therapeutic contexts and its potential applicability in human-robot interactions.
 
\begin{figure}[t!]
  \centering
  \includegraphics[width=8.0cm]{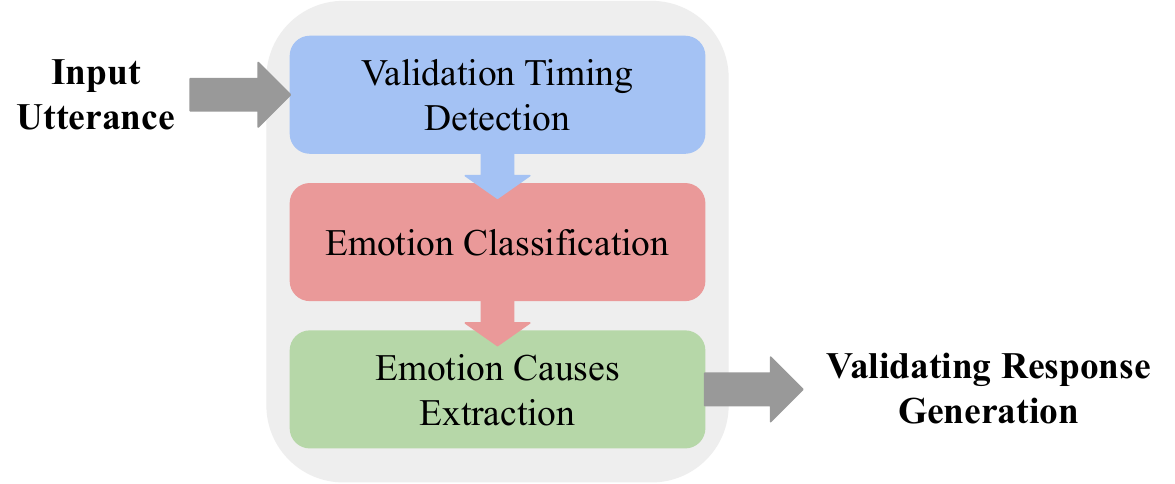} 
  \caption{Overall architecture of validating response generation system}
  \label{fig:architecture}
\end{figure}

In this research, we propose a novel framework designed for generating validating responses in dialogue systems. The framework's architecture, depicted in Fig. \ref{fig:architecture}, comprises three integral modules. The first module (validation timing detection) focuses on the detection of appropriate moments for generating validating responses, thereby identifying the timing when the system should engage in validation. The second module (users' emotional states identification) encompasses two subtasks: classification of the users' emotional types and discernment of the reasons underlying the emotions. The third and final module (validating response generation) pertains to the generation of validating responses, wherein the system constructs responses that acknowledge and affirm the users' emotional states. Each module plays a crucial role in the process of emotional validation: the validation timing detection module recognizes the users' emotional states and their need for validation; the emotional state identification module comprehends the nuances of users' emotions and the causative factors; and the validating response generation module focuses on expressing acknowledgment and acceptance of the users' emotions, reinforcing that their feelings are valid and understood.

\section{Dialogue Dataset}
\label{sec:2}

In this study, we primarily employed the Japanese EmpatheticDialogues \cite{sugiyama2023empirical} dataset, a Japanese text-based dialogue dataset encompassing interactions between two speakers. This dataset formed the cornerstone of our study, serving both as a training and evaluation set. Complementing this, to assess the applicability of our model in spoken dialogue scenarios, we utilized the TUT Emotional Storytelling Corpus (TESC) \cite{oishi2021design}. This speech-based dialogue dataset was instrumental in further evaluating the performance of our model in a much longer, spoken dialogue environment. Table \ref{tab:data_composition} shows the overall comparison of the two datasets, while the examples of dialogues from Japanese EmpatheticDialogues and TESC are shown in Table \ref{tab:example}.

\begin{table*}[h]
\caption{Specification of Japanese EmpatheticDialogues and TESC}
\label{tab:data_composition}
\centering
\begin{tabular}{lcccc}
\hline
Dataset & \#dialogue & \#utterance & Average \#word & Average \#turns \\
\hline
Japanese EmpatheticDialogues \cite{sugiyama2023empirical} & 20k & 80k & 23 & 4 \\
TESC \cite{oishi2021design} & 247 & 3080 & 41 & 16 \\
\hline
\end{tabular}
\end{table*}
\begin{table*}[]
\caption{Example of dialogues on Japanese EmpatheticDialogues and TESC}
\label{tab:example}
\centering
\begin{tabularx}{\textwidth}{llX}
\hline
\\
Japanese & & SPK1: この前のカラオケの時気付いたら門限すぎててさ \\ 
EmpatheticDialogues& & Last time at karaoke, I realized it was past my curfew.\\ 
& & SPK2: あんたのお母さんすごく厳しい人じゃなかった Isn't your mom really strict?\\
& &SPK1: うん着信履歴が10件以上あって見たとき手が震えたよ Yeah, my hands were shaking when I saw over ten missed calls from her.\\ 
& &SPK2: その気持ちわかるわー I totally understand how you feel. \\
\\
TESC & &SPK1: $\ldots$一回 大学夜ん時に帰ろうとしたんですけど ふと足元に違和感 ふと足元に違和感を感じて見てみたら このくらいの蛾みたいなのがいましてすんごいびっくりして すごい怖かった記憶 特に何が怖いって 彼ら いついきなりばって動いてくるかが分からない So, one night when I was about to head back from university, I suddenly felt something weird at my feet. I looked down and saw this huge moth, and it really freaked me out. The scariest part is not knowing when they'll suddenly start moving.\\ 
& &SPK2: そうですね Right.\\
& &SPK1: 分からないのが一番怖くて なんか予備動作があればいいんですけど It's scary not knowing when. I wish they had a warning sign.\\
& &SPK2: 突然動きだしますよ They do start moving all of a sudden.\\
& &$\ldots$ \\

\hline
\end{tabularx}
\end{table*}

\subsection{Japanese EmpatheticDialogues Dataset}
\label{subsec:2}

The Japanese EmpatheticDialogues \cite{sugiyama2023empirical} dataset was created after the original English EmpatheticDialogues \cite{rashkin2018towards}. The corpus comprises 20,000 dialogues, each consisting of four utterances exchanged alternately between a speaker and a listener, culminating in 80,000 utterance pairs. Originally, the dataset was characterized by 32 distinct emotion labels. However, due to the proximity and potential ambiguity of some labels, this study focuses on a refined subset. Adhering to Plutchik's wheel of emotions \cite{plutchik2001nature}, we have distilled the dataset to eight primary emotional states: fear, anger, surprise, disgust, sadness, joy, anticipation, and trust. This condensation was achieved by amalgamating closely related emotions, such as grouping `Terrified' and `Afraid' under fear, and so forth\footnote{Terrified \& Afraid \( \rightarrow \) Fear, Angry \& Furious \( \rightarrow \) Anger, Sad \& Sentimental \( \rightarrow \) Sadness, Excited \& Joyful \( \rightarrow \) Joy, and Hopeful \& Anticipating \( \rightarrow \) Anticipation}.

\subsection{TUT Emotional Storytelling Corpus (TESC)}

To evaluate our model performance in a spoken dialogue scenario, we utilized the TUT Emotional Storytelling Corpus (TESC) \cite{oishi2021design}, a Japanese multi-turn spoken dialogue dataset. This corpus encompasses interactions between student pairs who share a close bond. The experimental procedure involved one participant recounting a personal experience in response to an emotional prompt provided by the researcher. Concurrently, the listener engaged in active response, thereby maintaining a conversational environment reflective of everyday psychological interactions. TESC is categorized into the same eight emotional states as delineated by Plutchik's wheel of emotions \cite{plutchik2001nature}. The dataset comprises 247 conversational sessions involving 18 pairs of participants. Each session averages 133.9 seconds, culminating in a total of approximately 9.2 hours of dialogue. 


%
\section{Validation Timing Detection}
\label{sec:3}

This section delineates the initial module of our proposed system, commencing with an overview of the annotation process applied to the Japanese EmpatheticDialogues dataset and TESC for validation purposes. Following this, we introduce the validation timing detection model implemented in this study, culminating with a detailed presentation of the detection results.


\subsection{Annotation of Validation} %

In this subsection, we describe the process of annotating the Japanese EmpatheticDialogues dataset for the purpose of identifying the appropriate timing for generating validating responses. Our methodology involves classifying utterances into two distinct categories: those that warrant validating responses and those that do not. Initially, each utterance is coupled with its corresponding response. The determination of whether an utterance elicits a validating response is contingent upon the presence of specific validating phrases in the response \cite{pang2023prediction}, as identified through manual inspection and regular expression searches within the dataset. Key phrases indicative of validating responses include expressions like 「分かる(I understand how you feel)」, 「確かに (That is understandable)」, 「そう思う (I also think so)」, and 「それは+[感情言葉]+ね (That sounds [emotional word])」. Utterances devoid of these phrases are categorized as eliciting non-validating responses. For the purpose of this analysis, the input to all system modules comprises solely the utterance preceding the response. The dataset is subsequently segmented into training, validation, and testing subsets, following an 8:1:1 distribution ratio. To improve the model's predictive accuracy for the timing of valid responses, we expanded the dialogue history to include three previous utterances. This extension involves augmenting the data with the third utterance to enrich the existing dialogue context, adopting a \(A_1B_1A_2B_2\) format where `A' and `B' represent the speaker and listener, respectively, in chronological order. This enhanced dialogue history provides the model with a more comprehensive understanding of the conversational flow. Consequently, our analysis reveals that 29\% of the utterances (7110 in total) are classified as eliciting validating responses, with the remaining 17265 falling under the non-validating category.

Meanwhile, for the TESC corpus annotation, as it is a spoken dialogue-based dataset, the initial preprocessed step included the removal of backchannels, laughter, and filler utterances. Subsequent to this elimination, each remaining utterance was paired with its corresponding response. To maintain brevity, utterances were truncated to their final 50 words. The preprocessed utterances were then being annotated with the same step in Japanese EmpatheticDialogues dataset (except the data enhancement as the sequence of the spoken-dialogue dataset is not in form of \(A_1B_1A_2B_2\)). The final annotation results indicated that 260 utterances (approximately 17\%) fell into the validating response category, while 1280 were categorized under non-validating responses. The annotated data was subsequently divided into training, validation, and testing sets, adhering to a distribution ratio of 6:2:2. 




\subsection{Validation Timing Detection Model}
In this study, we employed the \textit{bert-large-japanese}\footnote{\url{https://huggingface.co/cl-tohoku/bert-large-japanese}} pre-trained model from Tohoku University, available on HuggingFace, as a foundational model for detecting validation timing in input utterances. Originally, this base model was pre-trained on Japanese Wikipedia articles, featuring paragraph-based text. This format diverges from our application domain of dialogue data. To our knowledge, there exists no pre-trained model specifically tailored for conversation dialogue-based data. To bridge this gap, we adopted a Task Adaptive Pre-Training (TAPT) \cite{gururangan2020dont} approach to enhance the model's performance for the validation timing detection task. We utilized the Japanese-Daily-Dialogue \cite{akama2023jdd} dataset, a resource rich in multi-turn daily conversation dialogues, to perform a masked-language-modelling (MLM) task on the BERT model. This adaptation of the model has been designated as \textbf{JDialogueBERT} for its specialized focus on dialogue. This step precedes the fine-tuning process on our target dataset, ensuring that the model acquires a comprehensive understanding of dialogue-based inputs, which is essential for our downstream task.

\subsection{Validation Timing Detection Result}\label{subsection:3.3}

In our study, the JDialogueBERT model underwent fine-tuning using the Japanese EmpatheticDialogues dataset, with hyperparameter optimization playing a pivotal role. Key parameters included a learning rate of 1e-05, a batch size of 64, 20 training epochs, and an evaluation every 100 steps focusing on precision with the Adam Optimizer. L2 normalization (weight decay rate of 0.01) and early stopping (patience threshold of 5) were implemented to mitigate overfitting.

Regarding evaluation metrics, the imbalanced data distribution in the dataset, where only 29\% of the data represents the target class, necessitated the use of macro-average precision, recall, and F1-score to assess model performance. We also specifically examined the precision, recall, and F1-score of the target class to evaluate its predictive accuracy in real-life conversation scenarios. For comparative analysis, we utilized a random baseline and the baseline BERT model. Additionally, we compared our model's performance with few-shot prompted ChatGPT\footnote{Prompt used:

``[Definition of validation stated in \ref{sec:1}]

Please classify each utterance into whether a validating response should be generated. Return validate if needed to generate a validating response and non-validate if not necessary to generate (meaning that it will generate a non-validating response)

[Followed by the two examples dialogues with validating response, and another two examples dialogues with non-validating response]''
} on the same task. 

The evaluation of our proposed model reveals its superior performance over comparative models, achieving a notable macro-average F1-score of 54.20\% and excelling in the target class with an F1-score of 43.14\%. This superiority is further underscored in its application to a spoken dialogue corpus, where it achieved a macro-average F1-score of 44.62\% and a target class F1-score of 27.36\%. It is important to note, however, that while ChatGPT demonstrated higher target class F1-scores in both datasets, predominantly due to its elevated recall values, this does not necessarily translate to greater real-world efficacy. In practical conversational scenarios, a model that frequently validates with high recall but low precision may fail to genuinely resonate with users, as it could give an impression of insincere understanding, diminishing the perceived empathy of the AI. Hence, despite ChatGPT's higher F1-scores driven by its recall, our model's superior precision makes it more apt for real-life conversational applications, providing responses that are more accurately aligned with the user's emotional context and content. The comprehensive results of our study are presented in Table \ref{tab:model_performance}, and Table \ref{tab:model_performance2}, respectively.

\begin{table*}[h]
\caption{Results of validation timing detection task on Japanese EmpatheticDialogues dataset [\%]}
\label{tab:model_performance}
\centering
\begin{tabular}{lccccccc}
\hline
\multicolumn{1}{c}{} & \multicolumn{3}{c}{Macro Average} & & \multicolumn{3}{c}{Target Class} \\
\cline{2-4}
\cline{6-8}
\multicolumn{1}{c}{} & Precision & Recall & F1-Score & & Precision & Recall & F1-Score \\ \hline
Random Baseline & 50.10 & 50.11 & 47.53 & & 29.58 & 51.92 & 37.69 \\
BERT & 54.30 & 55.17 & 52.07 & & 33.70 & 58.24 & 42.70 \\
ChatGPT & 53.97 & 50.58 & 26.15 & & 29.74 & \textbf{97.66} & \textbf{45.59}\\
JDialogueBERT (Ours) & \textbf{55.41} & \textbf{56.47} & \textbf{54.20} & & \textbf{35.28} & 55.49 & 43.14 \\
\hline
\end{tabular}
\end{table*}
\begin{table*}[h]
\caption{Results of validation timing detection task on TESC [\%]}
\label{tab:model_performance2}
\centering
\begin{tabular}{lccccccc}
\hline
\multicolumn{1}{c}{} & \multicolumn{3}{c}{Macro Average} & & \multicolumn{3}{c}{Target Class} \\
\cline{2-4}
\cline{6-8}
\multicolumn{1}{c}{} & Precision & Recall & F1-Score & & Precision & Recall & F1-Score \\ \hline
Random Baseline & 49.31  & 48.72 & 44.42 & & 14.81 & 41.67 & 21.86 \\
BERT & 49.35 & 48.77 & 42.55 & & 14.94 & 47.92 & 22.77 \\
ChatGPT & \textbf{58.25} & 23.27 & 20.30 & & 16.49 & \textbf{100.00} & \textbf{28.32}\\
JDialogueBERT (Ours) & 52.24 & \textbf{54.25} & \textbf{44.62} & & \textbf{17.68} & 60.42 & 27.36 \\
\hline
\end{tabular}
\end{table*}

\section{Users' Emotional States Identification}
\label{sec:4}

This section explores the identification of users' emotional states, a pivotal module of our system consisting of two key subtasks. The first subtask, emotion classification, focuses on identifying the types of emotions the users experience. The second, emotion causes extraction, is dedicated to understanding the reasons behind these emotions. 


\subsection{Emotion Classification}


We extends the application of the JDialogueBERT model, previously utilized for validation timing detection, to the task of emotion classification. The Task Adaptive Pre-Training (TAPT) model was fine-tuned to classify emotions based on Plutchik's eight-class wheel, utilizing the Japanese EmpatheticDialogues dataset and TESC corpus. Notably, the learning rate was adjusted to 3e-05, differing from the previous model settings in \ref{subsection:3.3}.

Evaluation metrics included macro-average precision, recall, F1-score, and accuracy. The model's performance was benchmarked against a random baseline, the standard BERT model, and ChatGPT with few-shot prompting\footnote{Prompt used:

``You are asked to classify the given conversation into one of the following eight emotions (Fear, Anger, Surprise, Disgust, Sadness, Joy, Anticipation, Trust).

[Followed by one example dialogues for each emotion label]''
}. A significant observation was ChatGPT's inability to classify approximately 33\% of Japanese EmpatheticDialogues and 7\% of TESC samples due to insufficient clarity in the emotional content of the utterances. Consequently, ChatGPT often assigned these utterances to either a neutral category or other emotion types, diverging from the intended classifications.

Our model demonstrated superior performance over all comparative models. In the Japanese EmpatheticDialogues, it achieved a macro-average F1-score of 76.88\% and an accuracy of 77.20\%. In the TESC corpus, it recorded a macro-average F1-score of 57.99\% and an accuracy of 58.77\%. The comprehensive results of our study are detailed in Table \ref{tab:emo_model_performance}, and Table \ref{tab:emo_model_performance2}, respectively. 

\begin{table*}[h]
\caption{Results of emotion classification task on Japanese EmpatheticDialogues dataset [\%]}
\label{tab:emo_model_performance}
\centering
\begin{tabular}{l c c c c}
\hline
\multicolumn{1}{c}{} & Precision & Recall & F1-Score & Accuracy \\ \hline
Random Baseline & 12.52 & 12.80 & 12.22  & 12.56 \\
BERT & 76.82 & 75.29 & 75.60  & 76.39 \\
ChatGPT & 68.12 & 60.51 & 62.40  & 61.81\\
JDialogueBERT (Ours) & \textbf{77.40} & \textbf{76.76} & \textbf{76.88}  & \textbf{77.20} \\
\hline
\end{tabular}
\end{table*}
\begin{table*}[h]
\caption{Results of emotion classification task on TESC [\%]}
\label{tab:emo_model_performance2}
\centering
\begin{tabular}{l c c c c}
\hline
\multicolumn{1}{c}{} & Precision & Recall & F1-Score & Accuracy \\ \hline
Random Baseline & 12.51 & 12.50 & 12.45  & 12.66 \\
BERT & 58.55 & 56.41 & 55.39  & 57.14 \\
ChatGPT & 60.83 & 52.43 & 52.59  & 52.28\\
JDialogueBERT (Ours) & \textbf{61.14} & \textbf{58.36} & \textbf{57.99}  & \textbf{58.77} \\
\hline
\end{tabular}
\end{table*}

\begin{table*}[]
\caption{Example of emotion cause annotation on Japanese EmpatheticDialogues}
\label{tab:ED_emo_cause_annotation}
\centering
\begin{tabularx}{\textwidth}{lXl}
\hline
Emotion & Dialogue & Emotion Cause \\
\hline
Joy & SPK1: 明日久しぶりに\textbf{ディズニーランド}に行くんだー。I'm going to \textbf{Disneyland} after a long time tomorrow! & \begin{tabular}[t]{@{}l@{}}ディズニーランド \\ Disneyland\end{tabular} \\
\hline
Disgust & SPK1: 幼稚園のママ友なのですが、ことあるごとにマウントを取ってくるのが面倒です。She's a mom friend from kindergarten, but it's bothersome how she always tries to one-up me at every opportunity.
 & \begin{tabular}[t]{@{}l@{}}アピール \\ Boast\end{tabular} \\
& SPK2: ママ友って、面倒なことが多そうですね。Mom friends seem to come with a lot of trouble.&  \\
& SPK1: 子どもの成長や夫の職業など、自分の家のほうがすごいって\textbf{アピール}が激しくて不愉快なんです。 She aggressively \textbf{boasts} about her child's development, her husband's job, and how her family is superior, which is really unpleasant.& \\   
\hline
\end{tabularx}
\end{table*}

\begin{table*}[]
\caption{Example of emotion cause annotation on TESC}
\label{tab:emo_cause_annotation}
\centering
\begin{tabularx}{\textwidth}{lXl}
\hline
Emotion & Dialogue & Emotion Cause \\
\hline
Trust & SPK1:じゃあ、やっぱりいい。はい、大丈夫です。やっぱりこう頼りになる人ていうのはいますよね。 Okay, it's still good. Yes, it's okay. I knew it. There are people who can be depended on. & 先生 Teacher (Mr.)       \\
& SPK2: いますね。Yes, there are. & \\
& SPK1: 世の中には、やっぱりですね。僕の指導教員のですね、OO\textbf{先生}はですね、ほんとに頼りになるんですね。There are people in the world, you know. \textbf{Mr.} OO, my advisor, is really dependable.  & \\
\hline
Surprise & SPK1: つむりながら、こうパサっやるとね。カサみたいな音がして、カサカサカサみたいな音が聞こえて。 When you pinch it, you can hear a cracking sound. It sounds like a rustling sound. & \begin{tabular}[t]{@{}l@{}}ゴキブリ\\ Cockroaches\end{tabular} \\ 
& SPK2: 最悪すね。That's the worst.&  \\
& SPK1: で、起きて急いでね電気点けてみたら、案の定\textbf{ゴキブリ}で、で、もう、ほんとに、驚いて分かんないけど。俺その時なんか二階で寝てるんだけど、二階から一階までダッシュで下りたんだけど、多分そん時人生の中で一番早く走った。小学三年生だけど、一番早く走ったていうのが。I woke up and rushed to turn on the light, and sure enough, there were \textbf{cockroaches}. I was sleeping on the second floor at the time, and I dashed down from the second floor to the first floor, probably running the fastest in my life. I was in the third grade, but it was the fastest time I had ever run. & \\     
\hline
\end{tabularx}
\end{table*}

\subsection{Emotion Cause Extraction}

This subsection addresses the emotion cause extraction subtask. It begins with an outline of the annotation process for emotion causes in both the Japanese EmpatheticDialogues dataset and TESC corpus, followed by the introduction of the model developed for emotion causes extraction. Finally, the results of this extraction process are presented and discussed.

\subsubsection{Emotion Cause Annotation}

As there was no annotation on emotion causes in the original Japanese EmpatheticDialogues dataset and TESC corpus, one of the authors undertook the meticulous task of annotating emotion causes for each dialogue. During the annotation process, the annotator was provided with the input utterances, along with the corresponding ground truth response and the identified ground truth emotion. The primary task for the annotator was to extract specific phrases from the original utterances that effectively represented the causes of the emotions conveyed in these utterances. Some example input utterance with annotated emotion causes is shown in Table \ref{tab:emo_cause_annotation}, and Table \ref{tab:ED_emo_cause_annotation}.


\subsubsection{Emotion Cause Extraction Model}







In conventional approaches, emotion cause extraction from contextual data typically relies on end-to-end models trained with extensive annotated datasets \cite{gao2021improving,li2019context}. However, our study faces a limitation due to the absence of such comprehensive datasets. In response to this challenge, we propose an innovative method for extracting emotion causes from input utterances, circumventing the need for additional model training. This method leverages the output of an existing emotion classification model to directly ascertain the causes of emotions. Our approach involves calculating an importance score for each token in relation to the predicted emotion, 
\(e\). This is achieved by backpropagating from the neuron corresponding to the predicted emotion and calculating the gradient of the embeddings, thus obtaining a weight for each input embedding token relative to the predicted emotion. The importance score for each token, \(i\), is then determined using the formula:

\begin{equation}
\text{score}(i) = E_{i} W_{ie}
\end{equation}

Here, 
\( E_{i} \) represents the embedding vector of the i-th token, and 
\( W_{ie} \) signifies the weight from the input embedding token to the predicted emotion. By evaluating these importance scores, we can identify which tokens, and thereby which segments of the input, are most influential in the model's emotion determination. These influential segments are posited as the emotional causes within the utterance, which is the central focus of our investigation.

\subsubsection{Emotion Cause Extraction Result}

To assess the efficacy of our proposed emotion cause extraction model, we conducted an evaluation comparing the top 3 extracted tokens with the annotated ground truth emotion cause phrases. A prediction was deemed correct if any of the extracted phrases matched the one in the ground truth. We calculated the accuracy using the entire test dataset. Additionally, as supplementary evaluation metrics, we computed the BERT Score (a BERT-based measure for text generation focusing on lexical semantic similarity between the generated response and ground truth) \cite{zhang2019bertscore} and the BLEU Score (evaluating the correspondence of the generated response to the ground truth) \cite{papineni2002bleu}.

For comparative analysis, we employed the same models used in the previous section, including a random baseline, baseline BERT, and few-shot prompted ChatGPT\footnote{Prompt used:

``You are asked to predict the emotion cause, in terms of phrases (with a maximum of 5 words), of the input utterance, and return the emotion cause in a string in Japanese only.

[Followed by three examples dialogues with its extracted emotion causes phrase]''
}, to extract emotion causes. Notably, ChatGPT often returned entire sentences rather than specific phrases. To ensure a fair comparison, we extracted the first five words generated by ChatGPT. Despite not calculating ChatGPT's accuracy due to its differing approach, our method demonstrated superior performance, achieving 73.00\% accuracy and a BERT Score of 61.44\%, as detailed in Table \ref{tab:emo_cause_result}.

\begin{table*}[h]
\caption{Result of emotion cause extraction task on Japanese EmpatheticDialogues dataset and TESC [\%]}
\label{tab:emo_cause_result}
\centering
\begin{tabular}{lccccccc}
\hline
\multicolumn{1}{c}{} & \multicolumn{3}{c}{Japanese EmpatheticDialogues} & & \multicolumn{3}{c}{TESC} \\
\cline{2-4}
\cline{6-8}
\multicolumn{1}{c}{} & Accuracy & BERT Score & BLEU Score & & Accuracy & BERT Score & BLEU Score \\ \hline
Random Baseline & 30.00 & 53.14 & 0.00 & & 27.08 & 54.50 & 1.35 \\
BERT & 68.00 & 59.94 & 0.77 & & \textbf{39.58} & \textbf{56.69} & 2.13 \\
ChatGPT &  & 54.91 & 0.15 & &  & 55.96 & \textbf{4.86}\\
JDialogueBERT (Ours) & \textbf{73.00} & \textbf{61.44} & \textbf{1.03} & & 33.33 & 56.10 & 2.21 \\
\hline
\end{tabular}
\end{table*}

However, the results on the TESC dataset, presented a less favorable outcome for our method compared to ChatGPT and baseline BERT. Our method, which incorporates task-adaptive pre-training on a dialogue dataset, might overly focus on dialogue-specific information, possibly obscuring more generalized context cues essential for emotion cause extraction. This specialized training could limit the model's ability to recognize broader contextual elements crucial in speech-based dialogues, as found in the TESC dataset. In contrast, BERT and ChatGPT, without undergoing additional dialogue-centric pre-training, may retain a broader understanding of context, facilitating more effective emotion cause extraction in such datasets.

\section{Validating Response Generation}
\label{sec:5}

This section examines the generation of validating responses by our proposed system. It commences with an introduction to the validating response generation model, followed by an evaluation of the model's performance.


\subsection{Validating Response Generation Model}

In our system, the generation of validating responses is achieved through a rule-based approach. When the initial module detects an input utterance as requiring a validating response, it predicts the emotion and potential emotion cause token using the second module. Based on the emotion cause and the predicted emotion category, the model generates a validating response. If the confidence level of the predicted emotion exceeds a threshold of 0.95, the model produces a response incorporating the emotional expression, formulated as 「確かに・分かる＋それは[感情言葉]ですね (That is understandable/I understand how you feel+That sounds [emotional words]」. If the confidence level is below this threshold, the response omits the emotional expression, resulting in a simpler 「確かに・分かる (That is understandable/I understand how you feel)」. Furthermore, when both the predicted emotion's confidence surpasses 0.95 and the identified emotion causes include nouns, the response is generated in the format of 「確かに・分かる＋[要因]は[感情言葉]ですね (That is understandable/I understand how you feel+[Reason] sounds [emotional words]」. This method ensures controlled generation of responses that are expected to support the emotional needs of the users. 



\subsection{Automatic Evaluation}

To assess the effectiveness of our validating response generation model, the BERT Score was selected as the primary evaluation metric. This involved computing the score between the generated response and the ground truth to ascertain their similarity. For comparative analysis, we chose ChatGPT\footnote{Prompt used:

``[Definition of validation stated in \ref{sec:1}]

Please generate a validating response for the given utterances. The generated response should be a validating response, with a maximum length of 15 characters, in Japanese.

[Followed by three examples dialogues with validating response]''
} and a standard Transformer-based Seq2Seq encoder-decoder generation model \cite{fu2023dual}. Our experimental findings indicate that our proposed method outperformed the comparative models, in both textual-based dialogue and spoken dialogue scenarios,  achieving a BERT Score of 59.34\% and 57.18\%, respectively. Detailed results of this evaluation are systematically presented in Table \ref{tab:generate_model_performance}. 

\begin{table}[h]
\caption{Objective evaluation (BERT Score) of validating response generation task on Japanese EmpatheticDialogues and TESC[\%]}
\label{tab:generate_model_performance}
\centering
\begin{tabular}{l c c}
\hline
\multicolumn{1}{c}{} & Japanese EmpatheticDialogues   & TESC  \\ \hline
Transformer \cite{fu2023dual} & 55.69 & 53.23 \\
ChatGPT & 58.20 &  57.02\\
JDialogueBERT (Ours) &  \textbf{59.34} & \textbf{57.18}  \\
\hline
\end{tabular}
\end{table}

\subsection{Human Evaluation}

To further assess the performance of validating response generation, we conducted an empirical A/B test against Transformer and ChatGPT. Thirty dialogues and their corresponding validating responses were randomly selected from each dataset. During the evaluation, participants were presented with two generated responses for each dialogue – one from JDialogueBERT and the other from either Transformer \cite{fu2023dual} or ChatGPT. Three annotators were tasked with determining the superior response based on criteria of naturalness, contextual understanding, and emotional understanding. Naturalness evaluated the human-like quality and grammatical accuracy of the response. Contextual understanding assessed the system's perceived grasp of the dialogue's context, while emotional understanding gauged the system's empathy and emotional resonance with the user's experience. Annotators were instructed to select the more effective response or declare a tie. 

The experimental results exhibit a significant preference for our method, with 47.8\% and 66.7\% of participants favoring our generated responses over those by Transformer in the Japanese EmpatheticDialogues and TESC, respectively. Moreover, when compared with ChatGPT, our method still maintained a higher preference rate, with 40.0\% and 48.9\% of participants in Japanese EmpatheticDialogues and TESC, respectively, opting for our generated responses. These findings underscore the effectiveness of our approach in generating more contextually and emotionally aligned responses in conversational AI systems. The comprehensive results of this evaluation are presented in Table \ref{tab:human_evaluation}.

On top of the human evaluation, we have conducted an additional comprehensive analysis focusing on the inter-annotator agreement, utilizing Cohen's Kappa \cite{mchugh2012interrater} to determine the inter-annotator reliability among three evaluators across two distinct datasets and models. The results, presented in Table \ref{tab:inter_agreement}, alongside comparative model outputs in Table \ref{tab:generation}, show a moderate agreement level among evaluators, with average kappa scores of 0.45, 0.43, and 0.50 for pairs 1/2, 1/3, and 2/3 respectively, indicating a consistent assessment framework.

A notable disparity emerged in agreement levels between text and speech-based datasets. For the text-based Japanese EmpatheticDialogues, the agreement was notably higher, with kappa scores of 0.50 and 0.60 when evaluators compared the performance against the Transformer and ChatGPT models, respectively. This higher level of agreement can be attributed to the inherent clarity and structured format of text-based data, which typically presents fewer ambiguities, thus facilitating more consistent evaluations. In contrast, the agreement levels were notably lower for the speech-based TESC dataset. Here, kappa scores were 0.36 for the Transformer and 0.37 for ChatGPT, falling into the `fair agreement' category. These lower agreement rates are likely attributable to the complexities inherent in speech data, including factors like users' preference of speaking style and longer context length in a single utterance turn, which introduce a higher degree of variability and subjectivity into the evaluation process.

\begin{table}[h]
\caption{Result of human A/B test on Japanese EmpatheticDialogues and TESC [\%]}
\label{tab:human_evaluation}
\setlength{\tabcolsep}{2mm}
\centering
\begin{tabular}{lccccccc}
\hline
 & \multicolumn{3}{c}{\begin{tabular}[c]{@{}c@{}}Japanese \\ EmpatheticDialogues\end{tabular}} & & \multicolumn{3}{c}{TESC} \\
\cline{2-4}
\cline{6-8}
\multicolumn{1}{c}{JDialogueBERT (ours) vs.} & Win & Loss & Tie & & Win & Loss & Tie \\ \hline
Transformer~\cite{fu2023dual} & 47.8 & 44.4 & 7.8 & & 66.7 & 25.6 & 7.8 \\
ChatGPT & 40.0 & 37.8 & 22.2 & & 48.9 & 38.9 & 12.2 \\
\hline
\end{tabular}
\end{table}
%

%
\begin{table}[h]
\caption{Result of inter-annotator agreement on Japanese EmpatheticDialogues and TESC [\%]}
\label{tab:inter_agreement}
\setlength{\tabcolsep}{2mm}
\centering
\begin{tabular}{lccccccc}
\hline
 & \multicolumn{2}{c}{\begin{tabular}[c]{@{}c@{}}Japanese \\ EmpatheticDialogues\end{tabular}} & & \multicolumn{2}{c}{TESC} & &\multirow{2}{*}{\begin{tabular}[c]{@{}c@{}}Average \\per Pairs\end{tabular}} \\
\cline{2-3}
\cline{5-6}
\multicolumn{1}{c}{Ours vs.} & Transformer & ChatGPT & & Transformer & ChatGPT & & \\ \hline
Pairs 1/2 & 0.57 & 0.59 & & 0.30 & 0.35 & & 0.45\\
Pairs 1/3 & 0.46 & 0.55 & & 0.41 & 0.29 & & 0.43\\
Pairs 2/3 & 0.48 & 0.64 & & 0.38 & 0.47 & & 0.50\\
\hline
\textit{Average} & 0.50 & 0.60 & & 0.36 & 0.37 & & 0.46\\
\hline
\end{tabular}
\end{table}
\begin{table*}[]
\caption{Case studies of our JDialogueBERT model and other models on the Japanese EmpatheticDialogues dataset and TESC}
\label{tab:generation}
\centering
\begin{tabularx}{\textwidth}{llX}
\hline
Example 1 & Context & SPK1: 家の近くで落雷があったみたい。I think there was a lightning strike near my house. \\
&& SPK2: 大きい音がしたの？ Was it loud?\\
&& SPK1: うん、今までで一番大きい音だったから本当に近かったみたい。めちゃくちゃびっくりしたよ！ Yeah, it was the loudest sound I've ever heard, and it was really close. I was so surprised! \\
& Transformer & SPK2: それはびっくりだね。That sounds surprising.  \\
& ChatGPT &  SPK2: 落雷、大丈夫だった？Lightning strike, were you okay?  \\
& JDialogueBERT (Ours) & \textbf{SPK2: 確かに！それはびっくりですね！That is understandable! That sounds surprising!} \\     

Example 2 & Context & SPK1:それじゃあ、僕が怖かったことなんですけど僕は虫がそこそこ苦手で。Well, something that scared me is that I'm somewhat afraid of bugs.  \\
& Transformer &  SPK2: それは怖いですね。That sounds scary.  \\
& ChatGPT &  SPK2: 虫は誰でも怖いことあるよ。Everyone can be scared of bugs at times. \\
& JDialogueBERT (Ours) & \textbf{SPK2: 虫が怖いですね！わかる！I understand how you feel! Bugs are scary!}  \\
\hline
\end{tabularx}
\end{table*}
%


\section{Conclusion}
\label{sec:6}

This study presents a novel system designed to generate validating responses, thereby enhancing empathetic dialogue. The system is composed of three key modules: 1) validation timing detection, 2) identification of users' emotional states, and 3) generation of validating responses. Employing a Task Adaptive Pre-Training (TAPT) approach with a BERT-based model, our method demonstrated superior performance across all modules compared to other models, including a random baseline, the baseline BERT, and ChatGPT, in both textual-based dialogue and spoken dialogue settings. As a direction for future research, we aim to conduct user experiments using the conversational robot \cite{inoue2020attentive}. This will enable us to evaluate our model's efficacy in complex, real-time conversational settings, further validating the utility of our proposed framework.

\begin{acknowledgement}

The authors would like to acknowledge Professor Mika Enomoto for providing us with access to the TUT Emotional Storytelling Corpus, which enabled us to analyze and draw conclusions from a vast amount of data. This work was also supported by KAKENHI (19H05691) and JST Moonshot R\&D Goal 1 Avatar Symbiotic Society Project (JPMJMS2011).

\end{acknowledgement}

\end{document}